\documentclass[11pt,a4paper,onecolumn]{article}
\usepackage{epsfig,subfigure, float}
\usepackage{cite}

\begin{document}

\title{Awareness and Self-Awareness \\for Multi-Robot Organisms}

\author{Serge Kernbach\\
\small Institute of Parallel and Distributed Systems, University of Stuttgart, \\
\small Universit\"atstr. 38, 70569 Stuttgart, Germany, \\
\small email: {\it serge.kernbach@ipvs.uni-stuttgart.de}}

\date{}
\maketitle

Awareness and self-awareness are two different notions related to knowing the environment and itself. In a general context, the mechanism of self-awareness belongs to a class of co-called "self-issues" (self-* or self-star): self-adaptation, self-repairing, self-replication, self-development or self-recovery. The self-* issues are connected in many ways to adaptability and evolvability, to the emergence of behavior and to the controllability of long-term developmental processes~\cite{KernbachHCR11}. Self-* are either natural properties of several systems, such as self-assembling of molecular networks, or may emerge as a result of homeostatic regulation. Different computational processes, leading to a global optimization, increasing scalability and reliability of collective systems, create such a homeostatic regulation. Moreover, conditions of ecological survival, imposed on such systems, lead to a discrimination between ``self'' and ``non-self'' as well as to the emergence of different self-phenomena. There are many profound challenges, such as understanding these mechanisms, or long-term predictability, which have a considerable impact on research in the area of artificial intelligence and intelligent systems.

The appearance of collective awareness in artificial social systems is another, very relevant, topic of modern research. Collective systems, such as swarms of insects, groups of animals or robots, or traffic systems possess several unique properties: scalability, reliability, adaptability to a large variation of environmental conditions. More generally, collective systems play very important role on Earth. We encounter them in all sizes, at all scales and in all forms, in biological and technological systems, in the oceans, in the air and on the ground. Basically, life, as we know it, is impossible without collective, such as multicellular or multi-individual, forms of existence. The mechanisms of awareness in such systems include several components: common knowledge, model of the environment, model of self, and reasoning with models, e.g. in the form of a planning process~\cite{Kornienko_S03A}. These collective mechanisms perform a very interesting task: the system models its environment and itself, and based on collective reasoning it recognizes itself (as the whole collective system) in the environment. The recognition of the collective self is comparable to the simplest forms of collective artificial preconsciousness, which is very hard to achieve, especially taking into account the distributed nature of collective systems.

Mechanisms of awareness and self-* properties are of especial interest in multicellular systems. Such systems consist of a large number of cells-modules, which can be connected to each other or behave independently like a swarm~\cite{Kernbach08_2}. Multicellular organisms are self-adaptive, self-regulative and self-developing, and are objects of research in such areas as artificial embryology or evolutionary computation, but are also of practical importance due to structural and functional reconfigurabilitry and adaptability. The SYMBRION and REPLICATOR projects\footnote{The SYMBRION project is funded by European Commission grant 216342 as part of the work programme Future and Emergent Technologies Proactive. The REPLICATOR project is funded as part of the work programme Cognitive Systems, Interaction, Robotics through grant 216240.} deal with artificial multicellular systems and different processes taking place in such systems~\cite{Levi10}. The main focus of these projects is to investigate and develop novel principles of adaptation and evolution for multi-robot organisms based on bio-inspired approaches and modern computing paradigms. Such robot organisms consist of a large-scale swarm of robots, which can dock with each other and symbiotically share energy and computational resources within a single artificial-life-form. In addition, the individual robots can be equipped with special tools and share information from remote or specific sensors. When it is advantageous to do so, these swarm robots can dynamically aggregate into one or many symbiotic organisms and collectively interact with the physical world via a variety of sensors and actuators.

\begin{figure}[htp]
\centering
\epsfig{file=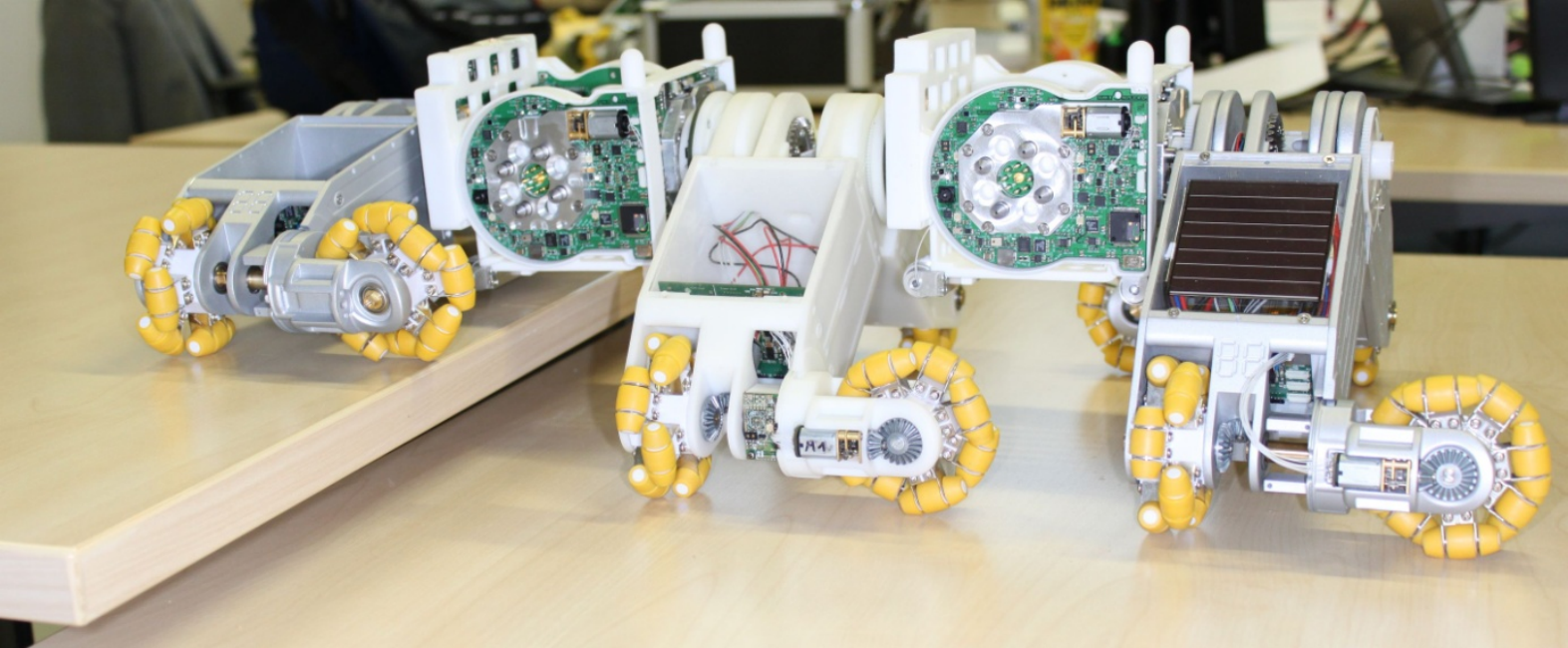,width=1.\textwidth}
\caption{Simple artificial organism consisting of five modules.\label{fig:organism}}
\end{figure}

Mechanisms of awareness and self-awareness in robot organisms are based on bio-inspired and evolutionary paradigms, such as artificial embryogenesis or on-line, on-board embodied evolution. For instance, the organisms are able to autonomously manage their own hardware and software organization, to reprogram themselves without human supervision. In this way, artificial robotic organisms become self-configuring, self-healing, self-optimizing and self-protecting from both hardware and software perspectives. The mechanisms of awareness and self-awareness can be either evolved, emerged through adaptive self-organization or appear as a result of homeostatic regulation.  This leads not only to extremely adaptive, evolve-able and scalable robotic systems, but also enables robot organisms and for new, previously unforeseen, functionality to emerge. The extraordinary potential and capability of autonomous large-scale self-aggregation, reprogramming and evolution would open-up a wide range of current and future applications. One of the main application scenarios of such artificial organisms is human-free environments with a high degree of danger or uncertainty.

\small
%\bibliographystyle{unsrt}
%\bibliography{../bibl_sk,../own_bibl_sk}

\end{document}